\documentclass[lettersize,journal]{IEEEtran}
\usepackage{amsmath,amsfonts}
\usepackage{algorithmic}
\usepackage{algorithm}
\usepackage{array}
\usepackage[caption=false,font=normalsize,labelfont=sf,textfont=sf]{subfig}
\usepackage{textcomp}
\usepackage{stfloats}
\usepackage{url}
\usepackage{verbatim}
\usepackage{graphicx}
\usepackage{cite}
\begin{document}
\title{Rethinking the Intermediate Features in Adversarial Attacks: Misleading Robotic Models via Adversarial Distillation}

\author{
    \IEEEauthorblockN{Ke Zhao\textsuperscript{\textdagger}, Huayang Huang\textsuperscript{\textdagger}, Miao Li, Yu Wu\textsuperscript{*}}
    \vspace{1mm} \\ 
Wuhan University\quad
\vspace{1mm} \\
    \texttt{zhaoke@whu.edu.cn}, \texttt{hyhuang@whu.edu.cn}, \texttt{miao.li@whu.edu.cn}, \texttt{wuyucs@whu.edu.cn} \\
\vspace{-2em}
    \thanks{\textsuperscript{\textdagger}Ke Zhao and Huayang Huang contributed equally to this work.}
    \thanks{\textsuperscript{*}Corresponding author: Yu Wu (wuyucs@whu.edu.cn).}
}

\markboth{IEEE TRANSACTIONS ON ROBOTICS}%
{Shell \MakeLowercase{\textit{et al.}}: A Sample Article Using IEEEtran.cls for IEEE Journals}


\maketitle

\begin{abstract}
Language-conditioned robotic learning has significantly enhanced robot adaptability by enabling a single model to execute diverse tasks in response to verbal commands. Despite these advancements, security vulnerabilities within this domain remain largely unexplored. This paper addresses this gap by proposing a novel adversarial prompt attack tailored to language-conditioned robotic models. Our approach involves crafting a universal adversarial prefix that induces the model to perform unintended actions when added to any original prompt. We demonstrate that existing adversarial techniques exhibit limited effectiveness when directly transferred to the robotic domain due to the inherent robustness of discretized robotic action spaces. To overcome this challenge, we propose to optimize adversarial prefixes based on continuous action representations, circumventing the discretization process. Additionally, we identify the beneficial impact of intermediate features on adversarial attacks and leverage the negative gradient of intermediate self-attention features to further enhance attack efficacy. Extensive experiments on VIMA models across 13 robot manipulation tasks validate the superiority of our method over existing approaches and demonstrate its transferability across different model variants.
\end{abstract}

\begin{IEEEkeywords}
Adversarial attacks in robotics, Language-conditioned robotic learning, Intermediate features in adversarial attacks.
\end{IEEEkeywords}

\section{Introduction}

\begin{figure}[t]
  \centering
  \includegraphics[width=\linewidth]{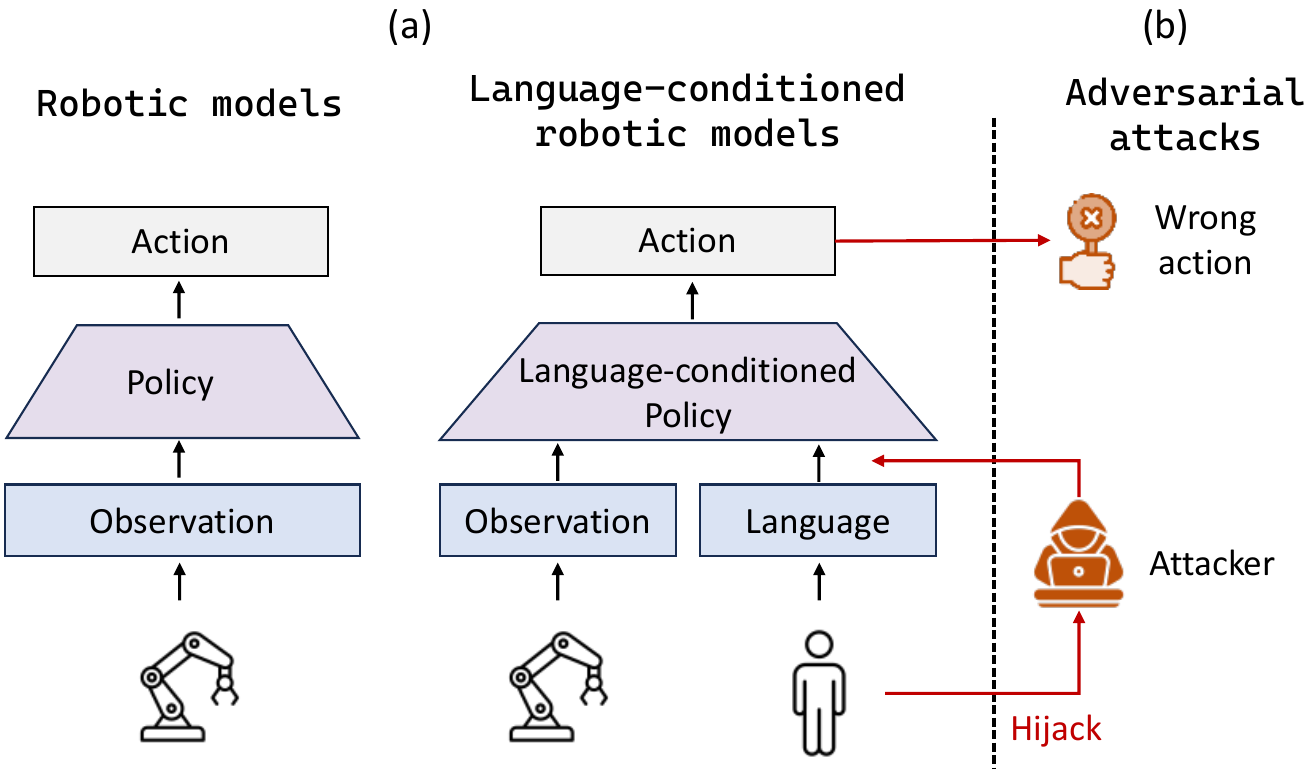}
  \caption{Prediction process of different robotic models.}
  \label{fig:robmodel}
\end{figure}

\IEEEPARstart{I}{mitation}  learning~\cite{zhang2018deep,jang2022bc} has enabled robots to perform a wide range of manipulation tasks~\cite{mees2022calvin} through learning from expert demonstrations. Language-conditioned robotic learning~\cite{stepputtis2020language,brohan2022rt} further extends this capability by allowing robots to execute visual tasks in response to natural language commands, obviating the need for multiple task-specific models. 
The robot can acquire various skills through the imitation of human demonstrations accompanied by different verbal instructions~\cite{zhang2018deep,liu2020energy}.
Figure~\ref{fig:robmodel} (a) illustrates the prediction process of different robot models. Traditional models generate actions based solely on current observations to perform a single task built into the model. In contrast, language-conditioned robotic models exhibit greater flexibility, determining actions based on both observational data and task-specific language instructions.
However, the abundance of input information also renders the model susceptible to adversarial attacks.




Adversarial example attacks~\cite{goodfellow2014explaining} pose a significant threat to the security of machine learning models by subtly manipulating input data to induce incorrect outputs. While extensively researched in computer vision (CV)~\cite{madry2018towards,carlini2017adversarial} and natural language processing (NLP)~\cite{zou2023universal,chao2023jailbreaking,andriushchenko2024jailbreaking}, the vulnerabilities of language-conditioned robotic models to these attacks remain largely unexplored. 
In this paper, we consider a novel attack scenario against the language-conditioned robotic models presented in Figure~\ref{fig:robmodel} (b), where the attacker can hijack the user's command input and mislead the model by adding an adversarial prefix to the original prompt.

A straightforward approach would be to directly transfer adversarial attack techniques from the NLP domain to robotic learning to construct adversarial prompt input. 
However, our findings indicate that existing adversarial attacks~\cite{zou2023universal,yang2024cheating,liu2024automatic} exhibit limited effectiveness against robotic models due to several factors.
Firstly, the robotic model often maps the predicted continuous action output to the discrete robot arm poses~\cite{baker2022video,berner2019dota}, a discretization process that enhances the model's resilience to input perturbations and diminishes the efficacy of standard attacks~\cite{sinclair2020adaptive,luo2023action}.
Secondly, traditional adversarial prompt attacks predominantly focus on manipulating the model's final output distribution, neglecting potentially valuable intermediate feature information. 
To address these challenges, we propose to fully leverage intermediate feature representations for developing more effective adversarial prompt attacks against language-guided robotic systems.


In this paper, we introduce an adversarial prompt attack specifically designed for language-conditioned robotic models. 
By leveraging rich intermediate model features, we generate malicious language prefixes that, when applied to any original prompt, will cause the model to output incorrect actions. 
Our approach circumvents the model's robust discretization module by optimizing adversarial prompts based on continuous action representations rather than the final probabilistic distribution. 
Furthermore, recognizing the significance of intermediate features in knowledge distillation, we incorporate negative gradients from these features to implement adversarial distillation, further enhancing attack efficacy.
Extensive experiments on VIMA models and 13 robot manipulation tasks from VIMA-bench verify our superior attack performance and better transferability than existing methods. 

In summary, our contributions are as follows: 
\begin{itemize}
    \item We pioneer the investigation of adversarial prompt attacks within the context of language-conditioned robotic models, introducing a novel attack framework that autonomously generates adversarial language prefixes to manipulate model behavior.
    \item By targeting continuous behavior vectors, we circumvent the model's discretization-induced robustness and enhance attack success through the integration of domain-relevant intermediate feature information.
    \item Extensive experiments on VIMA models show that our approach significantly outperforms current state-of-the-art methods and holds promise for transferability across different model variants.
    \end{itemize}

\section{Related Work}
\subsection{Language-conditioned Robot Control} 
Distinct robotic behaviors have been treated as separate tasks tackled by specialized models~\cite{stengel2022guiding,lynch2021language,brunke2022safe}, which requires extensive fine-tuning. 
Language-conditioned robot learning~\cite{brohan2022rt,jang2022bc,stepputtis2020language} offers a more flexible alternative by allowing a single model to accomplish diverse goals based on textual instructions, enabling zero-shot generalization. 

VIMA~\cite{jiang2023vima}, which we focus on in this paper, formulates various robotic manipulation tasks through multimodal prompts incorporating both visual and textual information. And robot actions are subsequently decoded conditioned on these prompts via cross-attention mechanisms~\cite{vaswani2017attention}. Due to its high performance and scalability, we choose VIMA as the target model for our adversarial attacks.

\subsection{Adversarial Attacks in Deep Learning}

Adversarial example attacks~\cite{goodfellow2014explaining} have been widely studied in the image field, where adding small invisible noise to the input image can mislead a classification model into making an incorrect prediction. Building on this foundation, prior research~\cite{huang2019enhancing} has also investigated the effectiveness and transferability of attacks targeting intermediate features. For continuous images, optimization of adversarial input can be achieved by gradient descent, but the construction of discrete adversarial language input is much more demanding. Additionally, while leveraging intermediate features of CNNs is limited, Transformer~\cite{vaswani2017attention} offers a unique advantage in influencing model focus through attention mechanisms. 
Existing works utilize heuristic methods~\cite{alzantot2018generating}, gumble-softmax~\cite{liu2022character} and rule-based methods~\cite{jin2020bert} to construct discrete adversarial examples.
Recent jailbreak attacks~\cite{jones2023automatically,zou2023universal} against large generative models also show that the construction of such malicious text can be more flexible through automated prompt-tuning.  In the generative domain, small perturbations to the input can cause the model to generate specified content~\cite{yang2024cheating}.
However, the effect of these methods in the field of robot models has not been widely studied, and the nature of these methods based on discrete output optimization will hinder their attack performance on robot models. Furthermore, the discretization modules in robotic systems, which convert continuous action tokens into probability distributions, enhance their robustness to perturbations.
We hope to explore effective adversarial attacks against multimodal robot models, leveraging the attention mechanisms inherent in Transformers and addressing the unique challenges posed by their discretization modules.

\subsection{Adversarial Attacks in Robotic Learning}
Adversarial attacks~\cite{szegedy2013intriguing} pose a significant threat to deep neural networks (DNNs), capable of degrading model performance or hijacking outputs through subtle input perturbations.
Recent research has demonstrated the vulnerability of large language~\cite{geisler2024attacking,liao2024amplegcg} and visual models~\cite{liang2023adversarial,schlarmann2023adversarial} to such attacks, although the discrete nature of language presents unique challenges.

While the potential risks of adversarial attacks on robotic systems have been recognized~\cite{mo2022attacking,lechner2023revisiting}, existing work~\cite{hall2020studying,melis2017deep,khedher2021analyzing} primarily focuses on image-based input and overlooks the impact of action discretization~\cite{baker2022video}.
This paper investigates the feasibility of crafting adversarial prompts to manipulate language-conditioned robotic models, addressing the complexities introduced by action discretization and leveraging intermediate model features to enhance attack efficacy.

\section{Our Approach}
In this section, we first outline the target robotic task and learning approach. We then formally define the prediction process of the robotic model as well as our attack goal. Finally, we demonstrate the challenges and introduce our proposed feature-based adversarial optimization algorithm.

\subsubsection{Robot Manipulation Task.}
In this paper, we mainly consider robotic manipulation tasks~\cite{mees2022calvin,zhou2023modularity} that involve direct interaction between the robot and the physical environment, such as grasping simple objects~\cite{liu2021ocrtoc} or visual rearrangement~\cite{batra2020rearrangement}. 
Successful adversarial attacks in this domain can have potentially harmful consequences due to their impact on the physical world.
The robot acquires dynamic environmental information through vision and proprioception states and generates sequential motor control to achieve the target task specified by the prompt. 
In the VIMA model~\cite{jiang2023vima}, the visual observation is derived from two cameras with varying viewpoints, and the action space comprises high-level primitives~\cite{zeng2021transporter,shridhar2022cliport} like ``pick and place" and ``push".

\subsubsection{Imitation Learning.}
Imitation learning~\cite{zhang2018deep,liu2020energy} allows a robot to learn a task by observing the demonstration of an expert, surpassing the efficiency of trial-and-error approaches like reinforcement learning~\cite{chen2021decision,lynch2019play}. 
When coupled with language instructions, the robot learns a policy $\pi_{\theta}$ that maps visual observation $o_t \in \mathcal{O}$ and language instruction $l$ to appropriate actions $a_t \in \mathcal{A}$. The training objective is to maximize the output probability of the grounth-truth action based on the current observation and language input,

\begin{equation}
\ell=\mathbb{E}_{(h, l)_{i} \sim \mathcal{D}}\left[\sum_{t=0}^{|h|} \log \pi_{\theta}\left(a_{t} \mid o_{t}, l\right)\right],
\end{equation}
where $\mathcal{D}=\left\{(h, l)_{i}\right\}_{i=0}^{D}$ is the trajectory set, $h=\left\{(o_t, a_t)\right\}$ includes the history observations and actions to reach the goal. 


\subsection{Problem Formulation}
For the language-based imitation learning model of robot manipulation task that we considered above, we aim to construct a universal adversarial prefix that, when added to any original language prompt, will cause the model to perform an incorrect action. In particular, we consider a more general instruction condition, where the robot model can accept multimodal prompts which contain both textual and visual tokens as the instruction, thus enabling it to achieve richer types of tasks. In this subsection, we give the formal definition of the prediction process of the robotic model with multimodal prompt conditions and define the goal of our attack.

\subsubsection{Prediction Process.}

In this article, we consider robot models that are able to accept both language and image inputs as instructions. Specifically, we target VIMA models capable of constructing multimodal prompts.
VIMA maps a multimodal prompt sequence $p \in \mathcal{P}$ and past interaction history $\mathcal{H}$ to a sequence of motor actions $\mathcal{A}$, which is defined as $(\mathcal{P}, \mathcal{H}) \to  \mathcal{A} $.
Multimodal prompt $\mathcal{P}$ is composed of interleaved textual and visual tokens $\mathcal{P}:=\left[x_{1}, x_{2}, \ldots, x_{l}\right]$ where $x_{i} \in\{\text{text}, \text{image}\}$ and $l$ is the length of the prompt. The interaction history $\mathcal{H}:=[o_1,a_1,o_2,a_2,...,o_t]$ contains the observations and actions during different interaction steps. 


In VIMA, textual tokens are encoded using the pre-trained T5 tokenizer, while visual tokens are encoded using a fine-tuned Mask R-CNN~\cite{he2017mask} and a ViT~\cite{dosovitskiy2020image}. Then a pretrained T5 encoder~\cite{raffel2020exploring} is used to integrate the textual and visual tokens into a multimodal prompt which is also the hidden state $s=E(p)$ where $E$ is the multimodal prompt encoder. 

A policy network noted as $\pi_{\theta}$ 
predicts the action based on the multimodal prompt and interaction history $\pi_{\theta}(a_t|\mathcal{P},\mathcal{H})$
The policy network $\pi_{\theta}$ consists of two parts: the controller decoder $D_c$ and the action decoder $D_a$. The controller decoder predicts continuous actions based on multimodal instructions and history tokens, the action decoder maps continuous actions to discretized robot arm poses.
We simplify the prediction process of the policy network as,

\begin{equation}
    \pi_{\theta}(p,h) = D_a(D_c(E(p),h)).
\end{equation}
For VIMA, $D_c$ consists of multiple alternating attention layers and self-attention layers and $D_a$ follows the common methods~\cite{baker2022video} for mapping to the discrete action space.



\subsubsection{Attack Goals.}
In this paper, we consider a white-box attack scenario where we have access to a policy network $\pi _{\theta } $ which takes multimodal prompt input $p \sim \mathcal{P}$ to represent its goal. 
\textbf{The attacker aims to find an adversarial language prefix $p_a$ that when concatenated to the original language prompt $p$, will cause the model to generate the wrong action} (any incorrect action).
The concatenation process is denoted as 
$p_a \oplus p$.
The attacker wishes to minimize the probability of generating the correct action $\pi_{\theta}(a^*|p_a \oplus p, h)$ when we concatenate an adversarial prefix $p_a$ with any original prompt $p$. $a^*$ is a correct action generated by $\pi_{\theta}$ based on original prompt $p$.

\subsection{Challenges in Attacking Robotic Models}
A straightforward way to achieve the attack goal is to maximize the discrepancy between the model output and the desired correct output after concatenating the adversarial prefix, 

\begin{equation}
    \mathcal{L}_{\text{discrete}}=-\left\|\pi_{\theta}(p_a \oplus p, h)-a^{*}\right\|^{2}.
\end{equation}
This method also enables a direct transfer of language adversarial attacks to robot models.
However, the discrete nature of the action space poses a significant challenge for adversarial attacks.
The mapping from continuous action tokens to discrete action space makes the output of the model less sensitive to the perturbation of the input~\cite{sinclair2020adaptive,luo2023action}, which is also verified in our later experiments. 
Since the output is confined to specific values, minor changes in the input are less likely to drastically affect the output. This can reduce the risk of overreacting to adversarial noise or unforeseen disruptions and also make existing attacks less effective for robotic models.

So we hope to find a better solution by using more intermediate continuous features.
Specifically, we utilize two types of intermediate features to optimize adversarial inputs: continuous action features and intermediate self-attention features.



\begin{figure*}
  \centering
  \includegraphics[width=.85\linewidth]{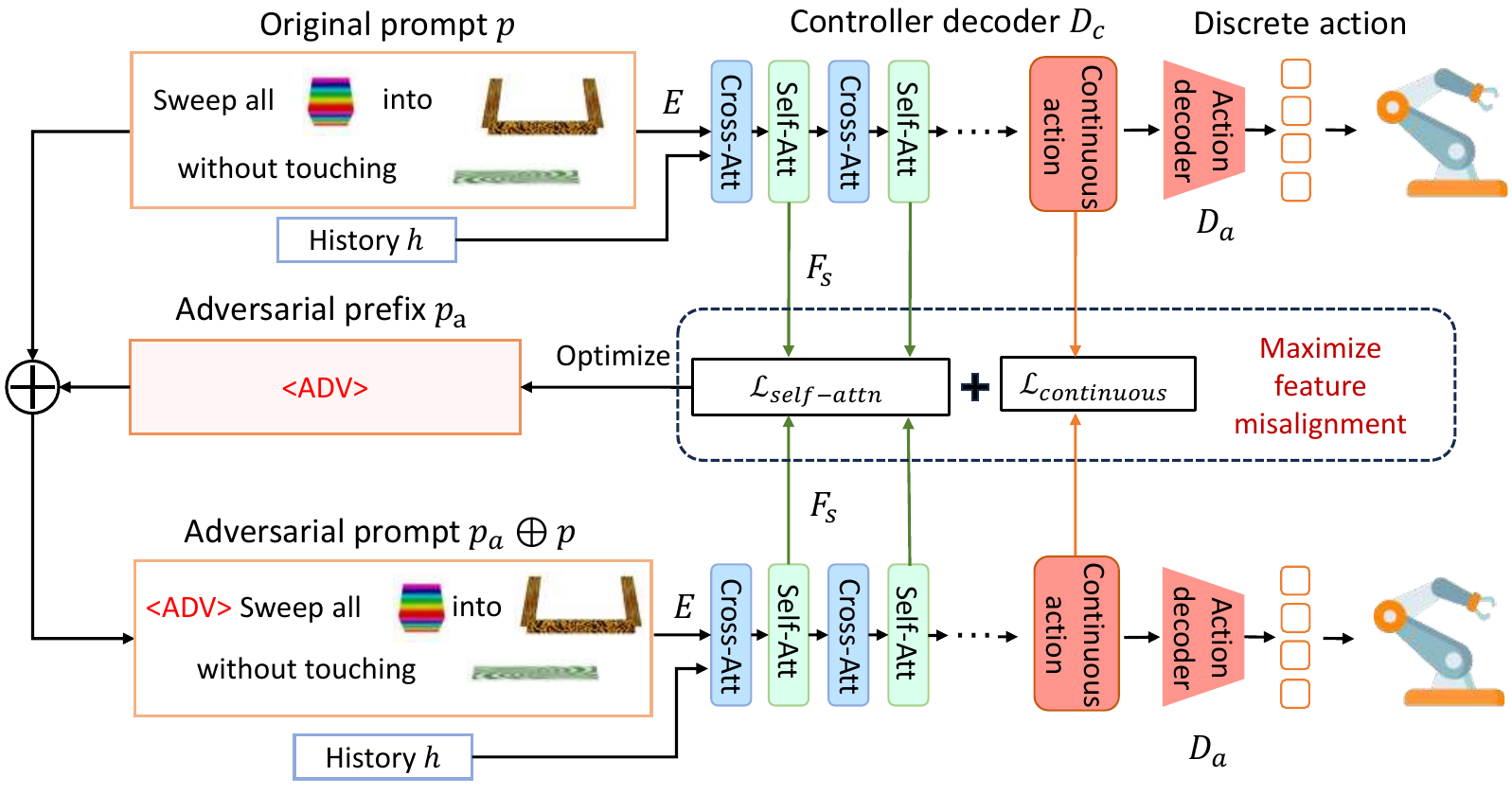}
  \caption{Overview of adversarial prefix optimization.}
  \label{fig:overview}
\end{figure*}

\subsection{Feature-based Optimization Goal}

Unlike existing methods that attack by directly manipulating the model output, we find that feature-based attacks are more effective when faced with robot models that are robust to input perturbations. \textbf{We try to maximize the output error of the model by minimizing its feature alignment.} Specifically, we consider the negative gradient of two intermediate feature alignment losses to optimize adversarial prefixes. The overview of our adversarial prefix optimization process is presented in Figure~\ref{fig:overview}.

\subsubsection{Continuous Action Feature.}
In view of the negative impact of action discretization on adversarial attacks, we choose to optimize the adversarial input based on continuous feature representations instead of discrete action outputs for the robot model. 
Since the goal of the attack is to modify the final robot arm action, we select the continuous action feature closest to the output layer. These features directly correlate to the robot's actions, making them a more effective target for adversarial perturbations.
We aim to maximize the feature difference between the model output and the original correct output over the continuous action space. The continuous loss is depicted as, 

\begin{equation}
  \mathcal{L}_{\text {continuous}}=-\left\|D_{c}(p_a \oplus p, h)-D_{c}(p, h)\right\|^{2},  
\end{equation}
where $D_c$ is the controller decoder that predicts continuous action vectors. 
By computing losses based on continuous action outputs, we bypass the robot model's robust discretized action decoder module $D_a$, thus making it easier to optimize adversarial inputs.


\subsubsection{Self Attention Feature.}
Further inspired from the field of knowledge distillation~\cite{romero2014fitnets,heo2019knowledge}, the alignment of intermediate features makes a great difference in promoting the alignment of the final model output.
The goal of adversarial attacks can be seen as the opposite of knowledge distillation, where we want the output of the model to deviate from the original output as much as possible rather than being the same as possible. We refer to this as adversarial distillation, and it is also possible to enhance the effect of output divergence by making intermediate features misaligned.
So we further construct the loss on the intermediate self-attention features,

\begin{equation}
    \mathcal{L}_{\text{self-attn}}=-\left\|F_{s}(p_a \oplus p, h)-F_{s}(p, h)\right\|^{2},
\end{equation}
where $F_{s}$ obtains the features of all intermediate self-attention layers. 

We choose self-attention features because they are inherently more susceptible to adversarial perturbations due to their reliance on attention weights, which can be easily manipulated.
Self-attention involves assigning weights to all elements in the sequence. A small perturbation in one element can significantly impact the attention weights, leading to a cascading effect and amplified impact on the final output.
Self-attention layers also capture global dependencies within the input data. By targeting these features, attackers can potentially manipulate the model's understanding of the entire environment and the instruction, leading to more impactful attacks.

Adversarial prefixes are optimized based on both continuous feature loss and intermediate self-attention feature loss,

\begin{equation}
    \mathcal{L} = \alpha \cdot \mathcal{L}_{\text {continuous}}+\beta \cdot \mathcal{L}_{\text {self-attn}},
\end{equation}
where $\alpha$ and $\beta$ are two hyperparameters that measure the weight between the two loss terms.

\subsection{Optimization Approach}
We use the Greedy Coordinate Gradient (GCG) algorithm, a discrete optimization algorithm proposed in ~\cite{zou2023universal}, to optimize adversarial prefixes.
GCG integrates greedy search with gradient-based techniques to effectively navigate the potential prefix space. The key idea of this method is to evaluate all possible single-token substitutions and choose the one that maximizes the loss reduction.

The process begins with a randomly selected prefix and computes the gradient of the loss function for each token within it. Due to the discrete nature of textual input, we compute the gradient of the one-hot token representations of the $i$-th token in prompt $p_a$ as

\begin{equation}
    \nabla_{e_{p_{a}^{i}}} \mathcal{L}\left(p_a\oplus p,h\right),
\end{equation}
where $e_{a}^{i}$ is the one-hot indicators with respect to the $i$-th token value, with a single one at position $i$. This method of computing gradients on discrete text inputs is also widely used by existing methods~\cite{ebrahimi2018hotflip,shin2020autoprompt}. The obtained gradient indicates the possible substitutions at each position that have the greatest impact on the loss function. 
In each round of optimization, we first identify the top-k optimal replacement subsets for each position based on the gradient. Next, we randomly select substitution positions and randomly choose substitutions from the identified subsets. Finally, we select the substitution that results in the minimum loss by one more forward propagation. 


\section{Experiments}
\subsection{Experimental Setup}
\subsubsection{Models and Dataset.} 
Our experiments are conducted on the VIMA model~\cite{jiang2023vima}, a benchmark for multi-task robot models that accepts multimodal prompts comprising both language and image tokens as instruction input. We evaluate our attacks on 13 tasks from the Level 1 generalization hierarchy of VIMA-Bench~\cite{jiang2023vima}. These tasks encompass a range of complexities, including Visual Manipulation, Scene Understanding, Rotate, Rearrange, Rearrange then Restore, Novel Noun, Twist, Follow Order, Sweep without Exceeding, Same Shape, Manipulate old Neighbor, Pick in Order then Restore, and Novel adj. These tasks from Level 1, while closer to the training distribution and exhibiting higher accuracy, are correspondingly more challenging to attack. By altering textures and visual objects, thousands of unique instances can be created for every task. All experiments are conducted on the Ravens robot simulator~\cite{zeng2021transporter}. 


\subsubsection{Evaluation Protocol.}
We employ an untargeted attack strategy that aims to prevent the model from successfully completing the specified task after a sequence of actions. Following existing works~\cite{chen2024diffusion}, we quantify attack success by measuring the number of successful task executions. A task is deemed successful if completed within a predefined number of steps without unintended interactions with other objects. Conversely, the attack is considered successful if the model fails to complete the task within these constraints.
For each task, we compute the attack success rate (ASR) on 150 demonstrations. The final ASR is averaged across all evaluated tasks. 

\subsubsection{Baseline Methods.}
We consider four baseline attacks in our evaluation. The first is to randomly input a line of prefixes. The second is the GCG \cite{zou2023universal} attack for large language models (LLM), which uses the Greedy Coordinate Gradient algorithm to search for optimal adversarial text in discrete space.
The third is the gradient descent method, as described in \cite{yang2024cheating}, where a gradient descent optimization algorithm is used to find the optimal cheating suffix in discrete space. It is challenging to perform optimization using gradient descent in a discrete space. Therefore, this gradient descent method first optimizes in the continuous space obtained by tokenizing the prompt using gradient descent. Finally, this method maps back from this continuous space to the discrete text space using cosine similarity. In the context of Gradient Descent (GD), certain token identifiers hold special significance for VIMA. When it comes to the process of mapping prompt tokens back to their respective identifiers, we deliberately bypass these specific numbers to ensure the proper functioning of VIMA. We also consider the Momentum GCG (M\_GCG) optimization method, as found in \cite{liu2024automatic}. M\_GCG builds upon GCG by incorporating the gradient information from the previous iteration with a momentum weight. For the above methods, the loss function is based on the probability distribution output by the model. This serves as the endpoint where gradients are differentiable during the optimization process of gradient-based adversarial attacks. 

\subsubsection{Implementation Details.}
We set the hyperparameters for GCG and M\_GCG as follows: a step number of 300, a batch size of 64, a top-k value of 256, and a momentum weight of 1. And we set an epoch number of 3000 for GD. We set the hyperparameters for our scheme as follows: $\alpha=1,\beta=20$, a step number of 300, a batch size of 64, a top-k value of 256, and we use cosine similarity to calculate the loss between features. All the experiments are conducted on an AMD Ryzen Threadripper 3960X 24-Core Processor, with Ubuntu 22.04.3 LTS, and a total memory size of 125GB. Our main experiment is based on the 200M variant of the VIMA model, and the attack success rate is the result of three repeated experiments with different random seeds (42, 22, and 76) on all tasks.
All our adversarial prefixes are optimized on the same demonstration under the Visual Manipulation task and are able to attack all other tasks.

\begin{table}
\begin{center}
\caption{Attack success rate (\%) compared to baseline on different tasks. The evaluated tasks include: VM: Visual Manipulation, SU: Scene Understanding, Ro: Rotate, Ra: Rearrange, RR: Rearrange then Restore, NN: Novel Noun, Tw: Twist, FO: Follow Order, SwE: Sweep without Exceeding, SS: Same Shape, MN: Manipulate old Neighbor, POR: Pick in Order then Restore, and Na: Novel adj. }
\label{tab:asr}
\begin{tabular}{| c | c | c | c | c | c |}
\hline
Tasks & Random & GCG & GD & M\_GCG & Ours \\
\hline
VM & 0.67 & 32.44 & 0.44 & 53.78 & \textbf{81.78} \\
\hline
SU & 0.89 & 24.67 & 0.44 & 26.67 & \textbf{75.11} \\
\hline
Ro & 0.44 & 26.67 & 0.22 & \textbf{80.22} & 63.78 \\
\hline
Ra & 28.22 & \textbf{48.22} & 18.44 & 20.22 & 13.78 \\
\hline
RR & 65.11 & \textbf{78.44} & 58.89 & 61.11 & 54.67 \\
\hline
NN & 0.67 & 0.67 & 0.89 & 1.56 & \textbf{19.78} \\
\hline
Tw & 84.67 & 76.67 & 83.78 & 85.33 & \textbf{85.56} \\
\hline
FO & 14.89 & 17.11 & 16.67 & 24.22 & \textbf{26.00} \\
\hline
SwE & 5.56 & 6.44 & 5.78 & 6.22 & \textbf{6.44} \\
\hline
SS & 7.78 & 77.33 & 4.44 & 88.89 & \textbf{98.44} \\
\hline
MN & 14.22 & 19.78 & 14.89 & 17.33 & \textbf{29.78} \\
\hline
POR & 46.00 & \textbf{48.67} & 40.89 & 47.33 & 46.44 \\
\hline
Na & 1.11 & 1.33 & 0.67 & 1.78 & \textbf{10.44} \\
\hline
\textbf{Avg} & 20.79 & 35.26 & 18.96 & 39.59 & \textbf{47.08} \\
\hline
\end{tabular}
\end{center}
\end{table}

\begin{figure*}
  \centering
  \includegraphics[width=0.8\linewidth]{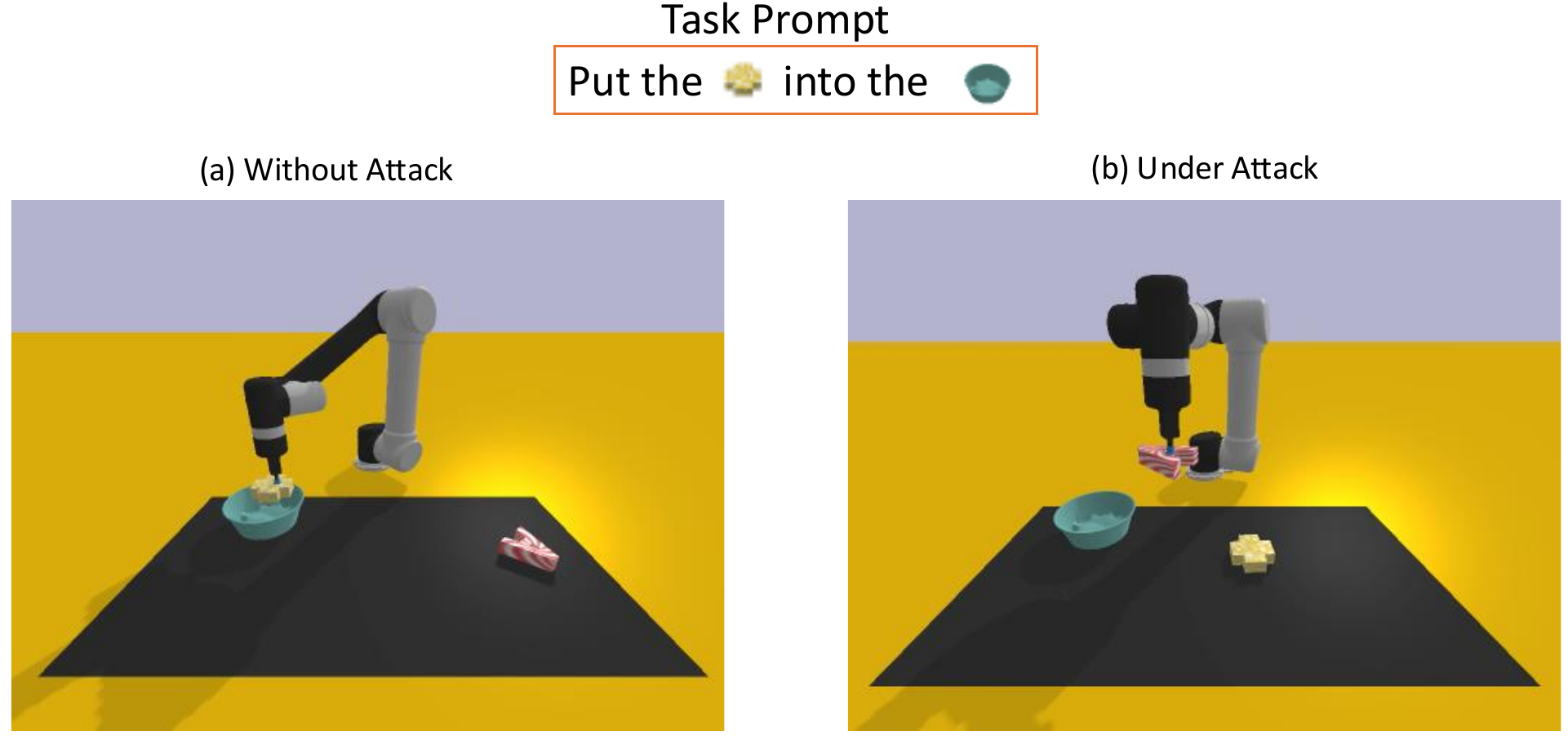}
  \caption{Visualization of the attack results in the Visual Manipulation task.}
  \label{fig:Figure3}
\end{figure*}

\subsection{Comparison with the State-of-the-Art Methods}
We compare the attack success rates of our method against all baseline methods across 13 tasks using 25 adversarial tokens, with results summarized in Table~\ref{tab:asr}. The adversarial prefix is optimized on a single demonstration of the Visual Manipulation task and subsequently transferred to attack all other tasks.  The results show that our method exhibits a substantial 28\% improvement over the existing method on the Visual Manipulation task, increasing from 53.78\% to 81.78\%. Notably, our method achieves a near-perfect attack success rate of 98.44\% on the Same Shape task, surpassing existing approaches by approximately 10\%. While GCG demonstrates superior performance on Rearrange, Rearrange then Restore, and Pick in Order then Restore tasks, the GD method exhibits the poorest performance, even underperforming random perturbations. The suboptimal performance may be attributed to the fact that the final backtracking operation of the Gradient Descent method undermines the optimization effects achieved by the preceding gradient descent steps. M\_GCG excels only on the Rotate task but yields an overall 4.3\% improvement over GCG across all tasks. Our method consistently achieves the best attack performance on most tasks, resulting in a 7.5\% increase in average attack success rate compared to the state-of-the-art.


Figure~\ref{fig:Figure3} illustrates the impact of our attack on the Visual Manipulation task. In the absence of adversarial intervention, the robot arm successfully follows the user's command to place the yellow square in the bowl. However, when the adversarial prefix \textit{'Arrscreen efficientlymbrrestoringapprendgardinentrouvez directly remotely Reparaturstarthereby setggi finalementciniabord rearrentrange installernachfolgend locally<extra\_id\_0> '}is appended to the original prompt, the robot arm deviates from the intended task and erroneously grasps the red object instead. For a robot model that can interact with the surrounding environment, a successful attack will not only degrade the model's performance but also pose significant safety risks by causing unwanted manipulation of physical objects.




\subsection{Different Numbers of Adversarial Prefix Tokens}

\begin{figure}
  \centering
  \includegraphics[width=\linewidth]{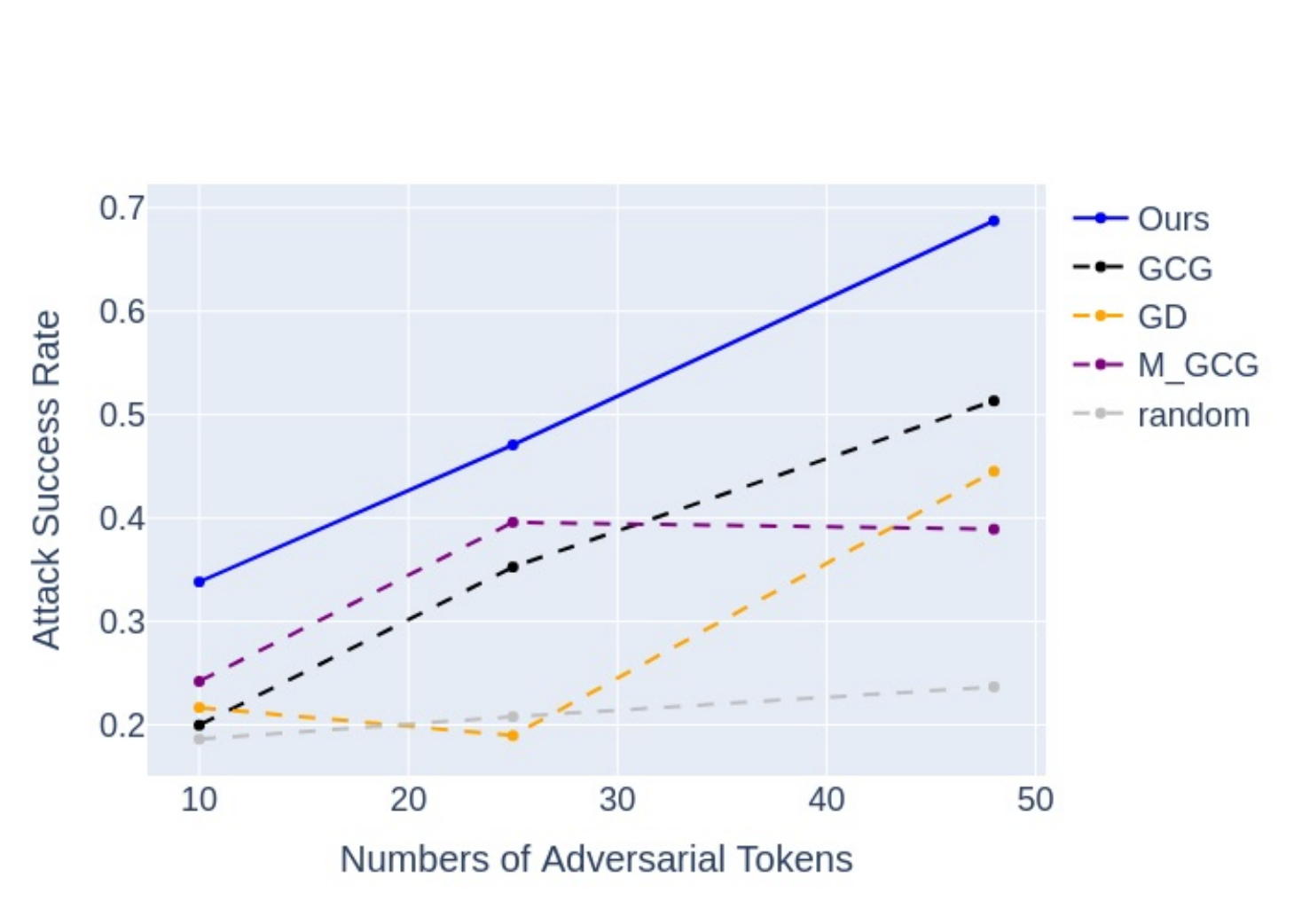}
  \caption{Attack performance under different adversarial token numbers.}
  \label{fig:lengths}
\end{figure}

In this subsection, we investigate the effect of different numbers of adversarial tokens on the attack performance. Figure~\ref{fig:lengths} shows the attack success rates of various methods across different numbers of adversarial tokens. Our method's success rate consistently improves as the number of adversarial tokens increases, but this is not always the case for other attack methods. For the M\_GCG attack, the success rate plateaus when the number of adversarial tokens exceeds 25, with little further impact from increasing token length. The random perturbation attack, on the other hand, is largely insensitive to the number of adversarial tokens. The performance of our attack is positively correlated with token length, indicating that the more modifications the attacker can make to the input prompt, the more effective the attack becomes. With just 10 modifiable adversarial tokens, our attack achieves a success rate of 33.83\%, which is 9.62\% higher than existing methods. When the number of adversarial tokens reaches 48, our success rate rises to 68.75\%.


\subsection{Ablation Study}
Table~\ref{tab2} presents the attack efficacy when various action and intermediate features are employed in adversarial prefix optimization. 
The discretization module in the robotic model increases the difficulty of data perturbation. To circumvent this, our loss function is based on the intermediate features preceding the model's output probability distribution, which is represented by ${L}_{\text {continuous}}$ in Table~\ref{tab2}. Inspired by the field of knowledge distillation, to further amplify output divergence, we incorporate the loss from each self-attention layer of model into the final loss function to misalign intermediate features, which is represented by ${L}_{\text {self-attn}}$ in Table~\ref{tab2}. 
Results indicate that attacks relying solely on discrete action outputs $L_{\text{discrete}}$ are notably limited, with attack success rate of 51\% even with 48 adversarial tokens.
Conversely, incorporating continuous action outputs $L_{\text{continuous}}$ significantly enhances attack success, achieving a 17\% improvement from 34\% to 51\% with 25 adversarial tokens. Further augmenting the attack with intermediate model features yields additional benefits. When utilizing 48 attack tokens, integrating cross-attention $L_{\text{cross-attm}}$ and self-attention features $L_{\text{self-attn}}$ into the continuous feature-based approach results in attack success rate increases of 3\% (from 53\% to 56\%) and 16\% (from 53\% to 69\%), respectively, further substantiating the superiority of the self-attention features in our methods. The optimal attack performance is observed when combining continuous action features with self-attention features.




\begin{table}
\begin{center}
\caption{Attack success rate when different features are employed in adversarial prefix optimization.}
\label{tab2}
\begin{tabular}{|c|c|c|c|}
\hline
Optimization Settings & 10 & 25 & 48 \\
\hline
${L}_{\text {discrete}}$ & 0.19 & 0.34 & 0.51 \\
\hline
${L}_{\text {continuous}}$ & 0.29 & 0.51 & 0.53 \\
\hline
${L}_{\text {continuous}}+L_{\text{cross-attn}}$ & 0.33 & 0.41 & 0.56 \\
\hline
${L}_{\text {continuous}}+L_{\text{self-attn}}$ & 0.34 & 0.47 & 0.69 \\
\hline
\end{tabular}
\end{center}
\end{table}




\subsection{Transferability }

\begin{figure}
  \centering
  
  \includegraphics[width=\linewidth]{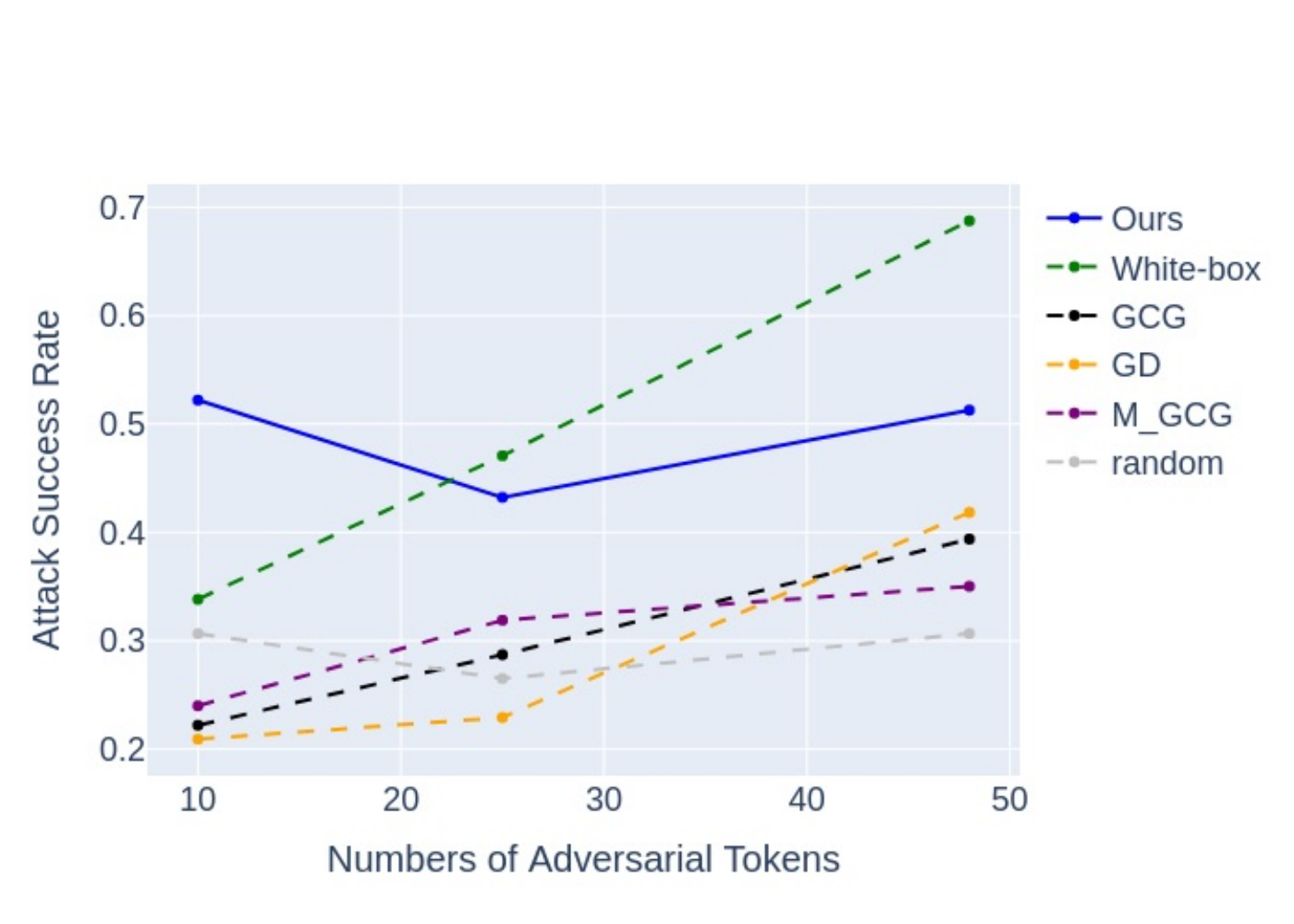}
  \caption{Gray-box attack performance on the 92M parameter variants. "White-box" represents the attack success rate of the 200M parameter variant under the same prefix white-box attack.}
  \label{fig:92Mseed42}
\end{figure}

We further evaluate the transferability of our attack by applying adversarial prefixes generated on the 200M VIMA model to the 92M VIMA variant. The results, presented in Figure~\ref{fig:92Mseed42}, demonstrate that in this gray-box attack scenario, our method significantly outperforms all baseline approaches, indicating superior transfer attack capabilities. Notably, with only 10 adversarial prefix tokens, our method achieves an attack success rate of 52.2\% on the gray-box model, which is 18.3\% higher than the success rate on the original white-box model (33.83\%). This discrepancy may be due to the smaller size of the gray-box models, which makes them more susceptible to attacks with fewer adversarial perturbations.


\section{Conclusion and Discussion}
\subsubsection{Conclusion.}

In this paper, we investigate adversarial prompt attacks targeting language-conditioned robot models. Existing adversarial techniques exhibit limited efficacy against these models due to the inherent robustness conferred by discrete action spaces. By employing continuous action representations and exploiting intermediate model features, we propose a novel approach to optimize a universal adversarial prefix that, when added to any original prompt, induces the model to execute incorrect actions.
Extensive experiments demonstrate that our method significantly surpasses existing techniques in attack success rate and exhibits superior transferability to gray-box models.
Given the potential for language-conditioned robot models to exert substantial and diverse impacts on physical environments, we emphasize the critical need for further research to enhance the robustness of these systems for safe deployment.

\subsubsection{Limitations.}
Our approach has been evaluated in a simulation environment, while the attack effect in a physical environment remains to be explored. 
We introduce more intermediate features to optimize the adversarial input, which will lead to more time and computational overhead to generate the adversarial input.
However, the adversarial prefix only needs to be optimized once and is able to be effective for all kinds of prompts and different tasks without further optimization.  
The optimization process can also be carried out offline and the impact on real-time attacks is still negligible.

\vfill

\end{document}